\documentclass[runningheads]{llncs}
\usepackage[T1]{fontenc}
\usepackage{graphicx}
\usepackage{tabularx}
\usepackage{diagbox}
\usepackage[hidelinks]{hyperref}
\usepackage{orcidlink}
\usepackage{bbding}  
\usepackage{fancyhdr} 
\begin{document}
\title{LLM-Enhanced Holonic Architecture for Self-Adaptive System of Systems}
\titlerunning{LLM-Ehnanced Holonic Architecture for Ad-Hoc Scalable SoS}
%
%
\author{Muhammad Ashfaq\inst{1}\textsuperscript{\Envelope}\orcidlink{0000-0003-1870-7680} \and Ahmed R. Sadik\inst{2}\orcidlink{0000-0001-8291-2211} \and Tommi Mikkonen\inst{1}\orcidlink{0000-0002-8540-9918} \and Muhammad Waseem\inst{1}\orcidlink{0000-0002-7994-3700} \and Niko M\"{a}kitalo\inst{1}\orcidlink{0000-0002-7994-3700}}
\authorrunning{Ashfaq et al.}
%
\institute{University of Jyv\"{a}skyl\"{a}, Jyv\"{a}skyl\"{a}, Finland
\email{\{muhammad.m.ashfaq, tommi.j.mikkonen, muhammad.m.waseem, niko.k.makitalo\}@jyu.fi}
\and
Honda Research Institute Europe, Germany\\
\email{ahmed.sadik@honda-ri.de}}
\maketitle              

\thispagestyle{fancy} 
\fancyhf{} 
\lhead{PREPRINT: This is a preprint of the paper accepted by Springer for publication in CCIS.} 
\pagestyle{plain} 

\setcounter{footnote}{0}  

\begin{abstract}
As modern system of systems (SoS) become increasingly adaptive and human-centred, traditional architectures often struggle to support interoperability, reconfigurability, and effective human-system interaction.
This paper addresses these challenges by advancing the state-of-the-art holonic architecture for SoS, offering two main contributions to support these adaptive needs.
First, we propose a layered architecture for holons, which includes reasoning, communication, and capabilities layers. This design facilitates seamless interoperability among heterogeneous constituent systems by improving data exchange and integration. 
Second, inspired by principles of intelligent manufacturing, we introduce \textit{specialised holons}--namely, supervisor, planner, task, and resource holons--aimed at enhancing the adaptability and reconfigurability of SoS.
These specialised holons utilise large language models within their reasoning layers to support decision-making and ensure real-time adaptability.
We demonstrate our approach through a 3D mobility case study focused on smart city transportation, showcasing its potential for managing complex, multimodal SoS environments.
Additionally, we propose evaluation methods to assess the architecture's efficiency and scalability, laying the groundwork for future empirical validations through simulations and real-world implementations.

\end{abstract}
\section{Introduction}
\label{sec:introduction}

A System of Systems (SoS) refers to an integrated collection of constituent systems (CS) that work together to provide capabilities that individual systems cannot achieve alone. SoS has revolutionised complex operations in critical areas, including military defence networks and healthcare systems~\cite{jamshidi2008systems}.
Advances in technology have created modern SoS that are dynamic, adaptive, and human-centred.
High dynamism allows an SoS to include, exclude, modify, or replace its CS during operation. This flexibility is essential for responding to changing mission requirements or improving system reliability and robustness during CS failures.
An adaptable SoS can operate effectively in uncertain and variable environments by interacting with the external environment.
Additionally, the SoS must support effective human interaction and decision-making throughout its operations.

These characteristics of modern SoS present new research challenges, which can be categorised into three main areas: interoperability, reconfigurability, and human interaction.
1) \textit{Interoperability}: Various entities develop, operate, and maintain their CS independently~\cite{lopes2016sos}. These CS often differ in their protocols, data formats, workflows, and interfaces~\cite{silva2020verification}. Ensuring interoperability among highly heterogeneous CS is a significant challenge, particularly for black-box CS--systems unknown at design time and discovered and integrated at runtime. Achieving interoperability facilitates knowledge sharing between CS, leading to better goal attainment for the SoS~\cite{song2022continuous,nadira2020towards}.
2) \textit{Reconfigurability}: An SoS must be capable of responding to inputs from the external environment, resulting in the reconfiguration at runtime~\cite{wudka2020reconfiguration,neto2017model}.
This reconfiguration needs to be sufficiently flexible to utilise the capabilities of black-box CS.
3) \textit{Human Interaction}: The SoS must support human-system interactions that enable informed decision-making without requiring extensive technical expertise from users.
For example, in vehicle platoons, each CS represents an autonomous car part of an urban mobility SoS~\cite{teixeira2020constituent}. Cooperative driving among various types of connected vehicles is crucial for tackling issues such as traffic congestion. To achieve this, the SoS must repeatedly reconfigure itself to create, change, or dissolve the platoon~\cite{wudka2020reconfiguration}.

The holonic architecture presents a promising framework by representing CS as self-governing entities, known as `holons'~\cite{blair2015holons}.
This approach utilises ontological descriptions to facilitate the discovery, dynamic composition, and runtime reasoning of CS~\cite{elhabbash2024principled}.
However, several areas require improvement.
First, creating ontological descriptions for CS relies heavily on manual input from vendors or system engineers. This dependency can be difficult to maintain in dynamic and large-scale environments.
Second, the architecture would benefit from more intuitive interaction mechanisms between humans and systems. At present, interactions demand technical expertise, which limits accessibility for non-expert users.


This paper is an extended version of our work published in ICSOFT~2024~\cite{icsoft24}, where we introduced the initial concept of integrating natural language processing (NLP) capabilities into the holonic architecture for SoS. 
However, this concept included a centralised NLP module, which was not conducive to autonomy and decentralisation.
In this extended version, we enhance the original framework in two significant ways. First, we propose a layered architecture for holons by introducing reasoning and communication layers. In this design, LLMs improve the interoperability of the holons. Second, drawing inspiration from intelligent manufacturing, we introduce \textit{specialised holons}--supervisor, planner, task, and resource--to enhance the adaptability and reconfigurability of SoS. Each holon is structured into three layers: 1) A \textit{reasoning layer} powered by LLMs for intelligent decision-making, 2) A \textit{ communication layer} built on the Robot Operating System (ROS) to invoke CS capabilities by sending specific commands. 3) A \textit{capabilities layer} that encapsulates the resources and capabilities of the CS. This combination improves holon-to-holon interoperability and human-SoS interaction while preserving the scalability advantages of the holonic architecture.

We present our approach through a case study that explores a 3D mobility holonic architecture for managing air and ground transportation in an urban environment. This case study demonstrates an SoS in which various transportation units, including ground and aerial vehicles, are modelled as holons. Additionally, we utilise LLM to facilitate natural language interactions between human operators and the systems. This implementation showcases how our extended architecture effectively addresses the key challenges of modern SoS while ensuring scalability and enhancing human-system interaction.

The remainder of this paper is organised as follows: The next section offers background information on the SoS, holonic architecture, and LLMs (Section~\ref{sec:background-and-related-work}). Following that, we will outline the proposed extended holonic architecture (Section~\ref{sec:enhanced-holonic-architecture}). We will then demonstrate this architecture through a case study focused on 3D mobility (Section~\ref{sec:case-study}). After that, we will discuss our findings and their implications (Section~\ref{sec:discussion}), along with potential implementation and evaluation methods (Section~\ref{sec:proposed-evaluation}). Finally, we will summarise our conclusions and highlight potential directions for future work (Section~\ref{sec:conclusion-and-future-work}).

\section{Background and Related Work}
\label{sec:background-and-related-work}

\subsection{System of Systems (SoS)}
\label{sec:system-of-systems}

An SoS is a collection of CS integrated into a larger framework, providing higher-level capabilities that individual CS cannot achieve alone~\cite{nielsen2015systems}.
A CS within an SoS exhibits unique characteristics that differentiate it from other systems, such as complex systems. These characteristics include:
\begin{itemize}
\item \textit{Autonomy}: CS are developed, operated, managed, and located independently while contributing to the overall objectives of the SoS~\cite{boardman2006system,maier1998architecting}.

\item \textit{Evolution}: The development of an SoS is an ongoing process. Its goals continually evolve, requiring the SoS to adapt to changing requirements, technological advancements, and environmental factors~\cite{maier1998architecting}.
This necessitates dynamic connectivity within the SoS based on mission needs rather than static configuration~\cite{boardman2006system}.
The evolution manifests in three key aspects: openness at the top for new high-level applications; openness at the bottom for technological upgrades; and continuous, gradual evolution that ensures operational stability~\cite{abbott2006open}.

\item \textit{Emergence}: The interactions among CS lead to higher-level capabilities that cannot be attributed to any single CS~\cite{maier1998architecting,boardman2006system}. The overall behaviour of the SoS can only be understood when integrated as a whole.
\end{itemize}

SoS are typically categorised into four types based on their management structure and goal alignment~\cite{dahmann2015systems,gideon2005taxonomy}.  Table~\ref{tab:sos-types} shows these types, where \textit{Semi} shows that CS are independent but SoS-level resources are centrally managed.



\begin{table}[ht]
\centering
\caption{Classification of SoS based on management structure and goal alignment.}
\label{tab:sos-types}
\begin{tabular}{|l|c|c|c|c|}
\hline
\diagbox[width=4cm]{\textbf{Characteristic}}{\textbf{Type}} & \textbf{Directed} & \textbf{Acknowledged} & \textbf{Collaborative} & \textbf{Virtual} \\
\hline
\textbf{Central management} & Yes & Semi & No & No\\
\hline
\textbf{Agreed-up goal} & Yes & Yes & Yes & No\\
\hline
\end{tabular}
\end{table}

Different architectural patterns are employed in SoS engineering, which can be categorised into three main types: centralised, hierarchical, and heterarchical.
The centralised architectures feature a single central controller with a top-down command structure. While they offer easy structure and global information sharing, they are generally unsuitable for complex SoS due to their single point of failure and limited scalability.
Hierarchical architectures employ multilayered control structures. They offer more flexibility and efficiency than centralised architectures and are suitable for directed and acknowledged SoS. However, they may face fault tolerance and scalability issues at lower levels.
Heterarchical architectures are fully decentralised and rely on the cooperation of autonomous CS. They offer high flexibility and eliminate single points of failure, making them suitable for collaborative and virtual SoS. However, they may struggle to achieve global optimisation, potentially leading to unpredictable behaviour.

\subsection{Holonic Architecture}
\label{sec:holonic-architecture}

Holonic architecture is inspired by Arthur Koestler's idea of holons and holarchies in biological and social systems~\cite{koestler1967ghost}. Holons, derived from the Greek words ``holos'' (whole) and ``on'' (part), are semi-autonomous entities that operate both independently and as part of a larger whole. These holons are organised into a holarchy, which is a hierarchical structure of self-similar entities. This structure facilitates both top-down decomposition and bottom-up composition, creating a flexible and adaptive framework for system design. Holonic architecture has found applications in smart manufacturing by decentralising control and enhancing resilience through autonomous components~\cite{sadik2017combining}.
Each holon has specific characteristics, including autonomy, social responsiveness, proactivity, and varying responsibilities based on its type.

Due to their dual nature and expressiveness, holons are particularly effective for modelling CS, as they accurately represent individual functions and contributions to an overall SoS~\cite{blair2015holons}. Nundloll~et~al.~\cite{nundloll2020ontological} explored the application of holonic principles in Internet of Things (IoT) system modelling and presented a framework that represents holons through ontologies. Elhabbash~et~al.~\cite{elhabbash2024principled} adapted this framework to the SoS domain, proposing an architecture where CS function as ontological holons. This design enables these CS to reason and communicate with one another, facilitating CS discovery and dynamic composition of the SoS.
Zhang~et~al.~\cite{zhang2023nlp} developed an automated approach using NLP techniques to derive holon ontologies from web data, specifically focusing on IoT devices rather than SoS.
Sadik~et~al.~\cite{sadik2023self} utilised holonic architecture to address the scalability issues in SoS by allowing runtime transitions between centralised, hierarchical, and holonic architecture patterns based on system conditions.
However, this architecture currently lacks mechanisms for effective human-to-holon interactions and communication with unknown holons. 

\subsection{Large Language Models (LLMs)}
\label{sec:large-language-models}

NLP has become a fundamental aspect of artificial intelligence, enabling more sophisticated human-computer interactions~\cite{khurana2023natural}. This field includes various computational techniques designed to understand, interpret, and generate human language. Among these techniques, LLMs are built on transformer architecture, a specialised neural network design that utilises self-attention mechanisms and parallel processing capabilities~\cite{zhao2023survey}.

Trained on vast amounts of data, LLMs develop advanced linguistic skills such as pattern recognition, structural analysis, contextual understanding, and semantic interpretation. These capabilities enable LLMs to perform complex tasks like text classification, sentiment analysis, and machine translation~\cite{radford2019language,brown2020language}.

LLMs have demonstrated significant value in software development, particularly in programming assistance and code generation~\cite{sadik2023analysis}. Recent research has explored integrating LLMs into human-robot collaboration~\cite{koubaa2024next}. Building on these findings, this study aims to investigate the integration of LLMs into SoS and holonic architecture.


\section{Enhanced Holonic Architecture}
\label{sec:enhanced-holonic-architecture}

\subsection{Holon}
\label{sec:holon}
\begin{figure}[t]
\centering
\includegraphics[width=\linewidth]{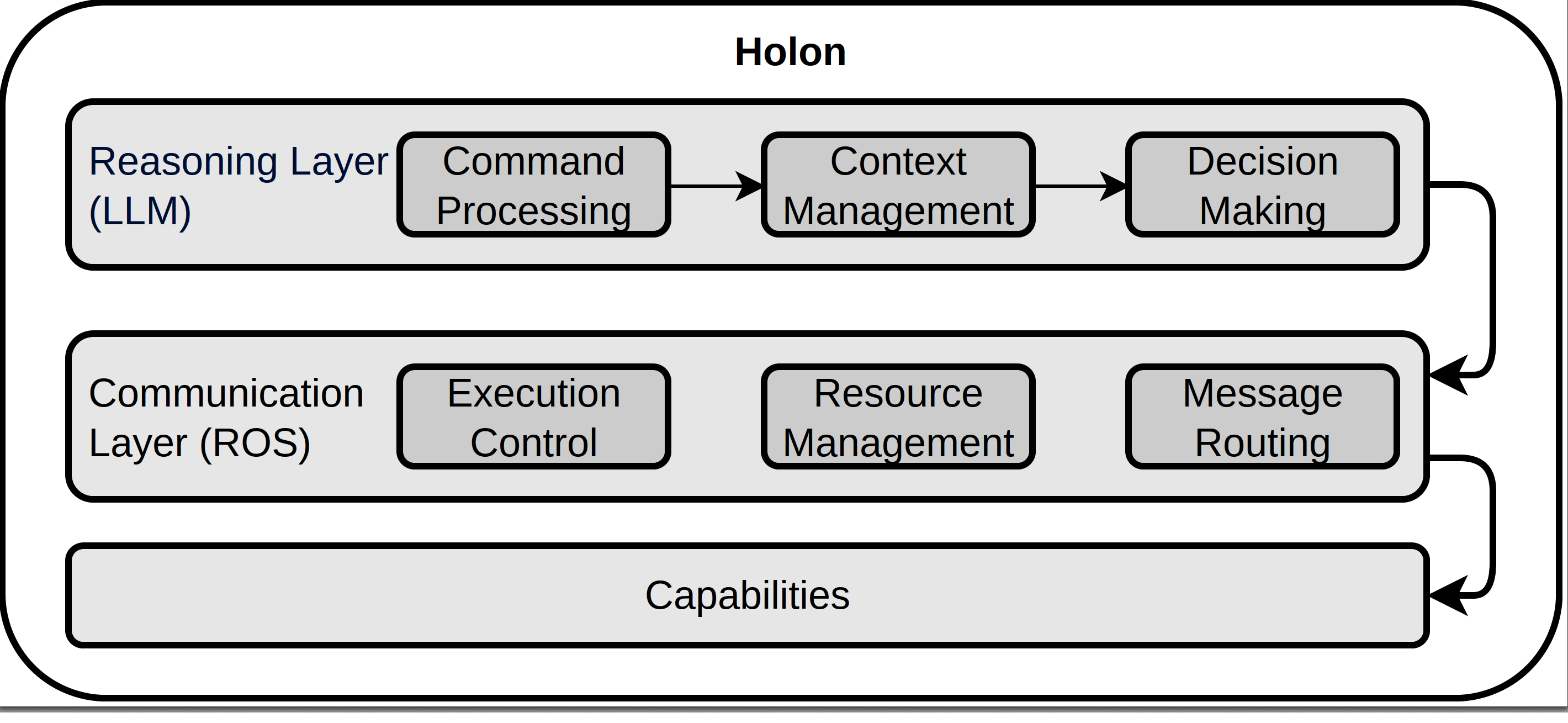}
\caption{Layered structure of a holon showing the hierarchical relationship between reasoning, communication, and capabilities layers, with their respective sub-components and interactions.}
\label{fig:holon}
\end{figure}

The holon is the core component of the proposed architecture, encapsulating the CS of an SoS. Each holon consists of three layers: reasoning, communication, and capabilities, as illustrated in Fig.~\ref{fig:holon}.

\subsubsection{Reasoning Layer:}
This layer enables context-aware decision-making and supports dynamic task planning and adaptation. The LLM is the central reasoning core, interpreting commands and facilitating complex reasoning across all holons. The reasoning layer comprises three sub-components:

\begin{enumerate}
\item \textit{Command Processing:} Preprocesses raw external inputs.
\item \textit{Context Management:} Crafts precise and contextually relevant prompts to ensure inputs are accurately tailored to the holon type, enhancing the effectiveness and precision of the LLM's responses. One implementation strategy involves using domain-specific ontologies, as demonstrated by Koubaa et al.~\cite{koubaa2024next}.
\item \textit{Decision Making:} Processes refined commands, creates appropriate action plans, and converts them into syntax comprehensible to the communication layer.
\end{enumerate}

\subsubsection{Communication Layer:}
This layer translates LLM instructions into concrete and actionable commands for the holon, enabling capability control. It consists of three sub-components:
\begin{enumerate}
\item \textit{Execution Control:} Implements instructions from the reasoning layer through the Robot Operating System (ROS), controlling the holon's operational capabilities.
\item \textit{Resource Management:} Monitors and allocates available resources for task performance.
\item \textit{Message Routing:} Facilitates real-time data flow and system description exchange between holons for coordination.
\end{enumerate}

\subsubsection{Capabilities Layer:}
Each holon possesses specific resources and services that are represented as integrated capabilities.

\subsection{Specialized Holons}
\label{sec:specialized-holons}

\begin{figure}[t]
\centering
\includegraphics[width=\linewidth]{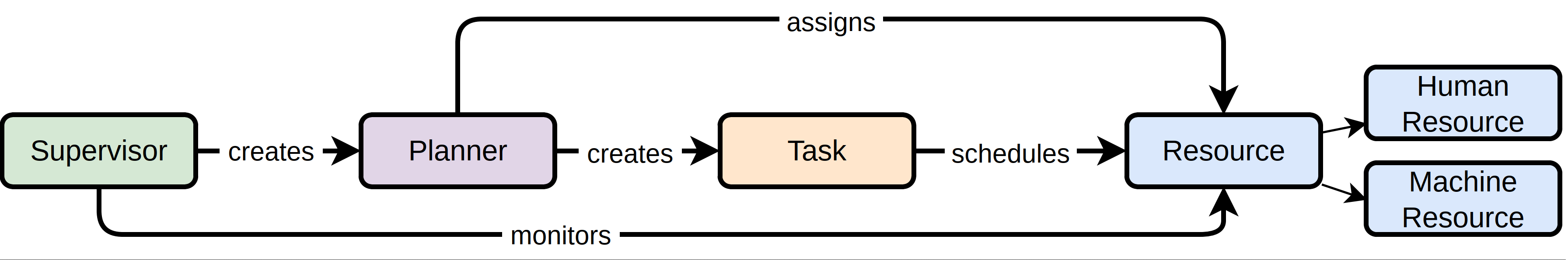}
\caption{Specialised holons and the workflow relationships among them.}
\label{fig:specialised-holons}
\end{figure}

Our architecture incorporates specialised holons designed to perform specific tasks that support SoS missions. Each holon employs an LLM tailored to its designated role, providing role-specific contextual information to the incoming commands. Figure~\ref{fig:specialised-holons} shows these holons and their workflows.

\subsubsection{Supervisor Holon:}
This holon coordinates the communication and schedule plans between the SoS and CS. It generates strategic and operational plans across the SoS while monitoring resource availability and utilisation, coordinating timing, and managing task interactions.

\subsubsection{Planner Holon:}
This holon translates the broad objectives of the supervisor holon into specific tasks. It develops detailed task plans from high-level goals, maintains a comprehensive overview of current tasks and resources, and optimises resource allocation across plans.

\subsubsection{Task Holon:}
This holon oversees individual task execution with real-time adaptability. It schedules and implements specific tasks, processes real-time environmental feedback, and adjusts task execution based on real-time conditions.

\subsubsection{Resource Holon:}
This holon exists in two specialised forms, each interfacing with the LLM for distinct purposes:
\begin{enumerate}
\item \textit{Human Resource Holon:} Encapsulates humans as holistic entities. It includes a human-machine interface that serves as the interface between humans and the SoS.
\item \textit{Machine Resource Holon:} Encapsulates the physical entities within the SoS. It ensures optimal resource utilisation based on task requirements and maintains real-time awareness of all sensors and modalities of the machine resources.
\end{enumerate}

\section{Case Study: 3D Mobility SoS Holonic Architecture}
\label{sec:case-study}

\subsection{Overview and Challenges}
To demonstrate our enhanced holonic architecture, we apply it to a 3D mobility case study. This case study represents a complex SoS that integrates unmanned ground vehicles (UGVs) and unmanned aerial vehicles (UAVs) in an urban environment~\cite{tchappi2020critical}. The 3D mobility SoS must coordinate with various unmanned vehicles, infrastructure elements, and human operators to provide efficient and safe transportation services.

Consider a scenario where a smart city customer requires transportation from point $X$ to point $Y$.
Customers may have diverse goals, such as choosing the fastest or least expensive route. This scenario presents several challenges.

\begin{itemize}
\item \textit{Complex Urban Environment}: Navigating through a densely populated urban area with varying altitudes and restricted flight zones for UAVs.
\item \textit{Dynamic Routing}: Adapting to traffic and environmental conditions in real-time to ensure the best route based on the customer's requirements.
\item \textit{Vehicle Coordination}: Seamlessly transitioning between UAVs and UGVs while maintaining a consistent and comfortable experience.
\item \textit{communication}: Ensuring clear and efficient communication between the user, vehicles, and the control centre to manage expectations and adapt to any changes in the mission.
\end{itemize}

\subsection{Realisation of the Scenario in Extended Holonic Architecture}
\begin{figure}[ht!]
\centering
\includegraphics[width=\linewidth]{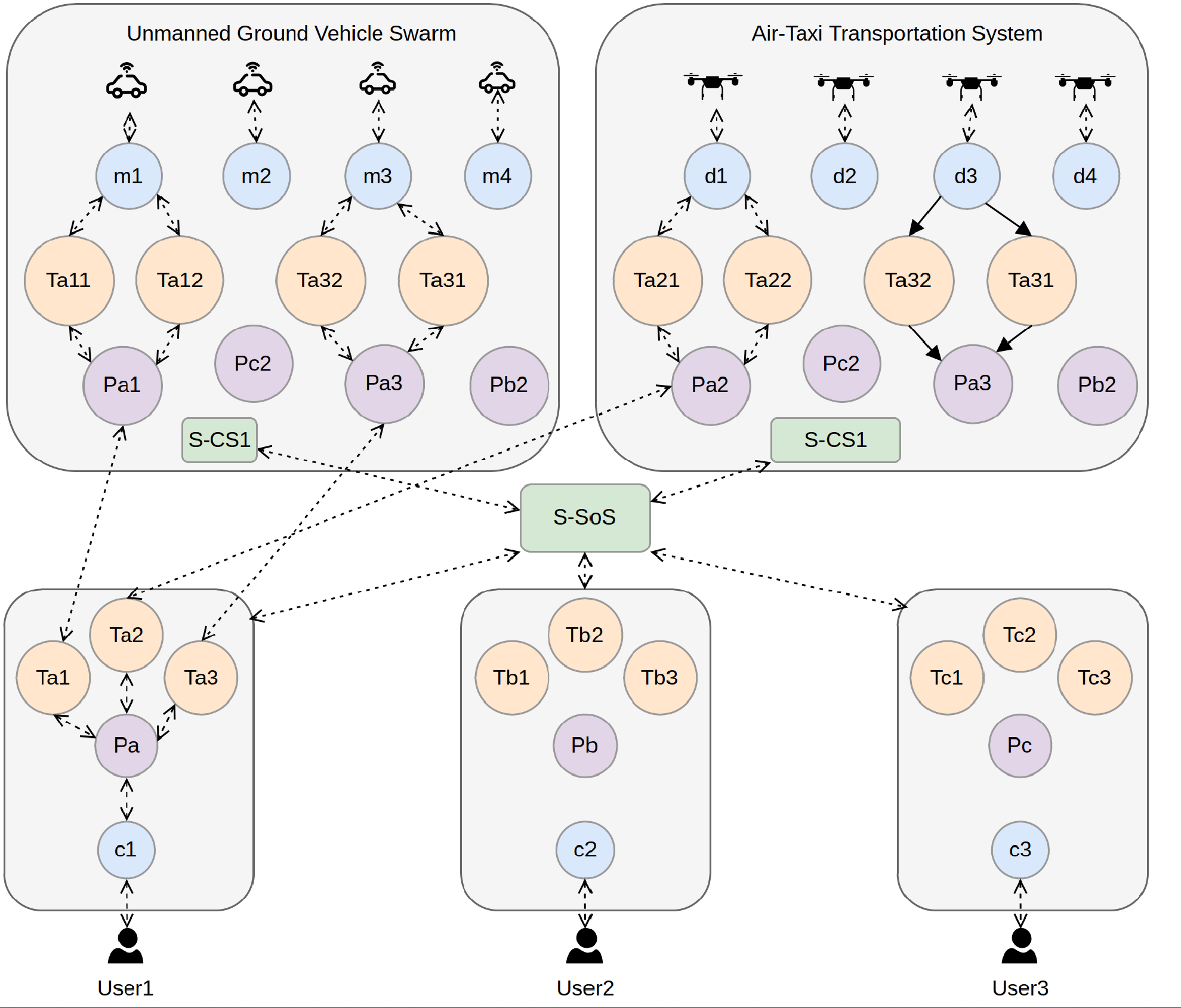}
\caption{3D Mobility Holonic Architecture for coordinating air and ground transport resources in urban environments.}
\label{fig:3d-mobility-holonic-architecture}
\end{figure}

\begin{figure}[ht!]
\centering
\includegraphics[width=\linewidth]{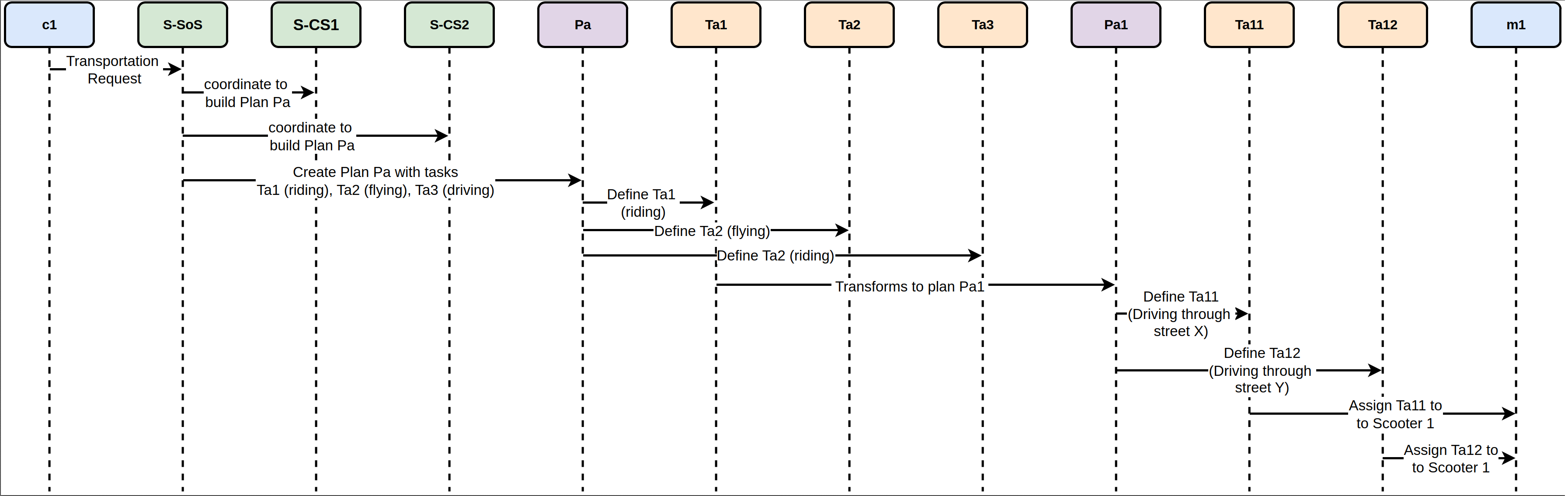}
\caption{Sequence diagram of the 3D Mobility SoS, illustrating task coordination from user request to plan execution across transport modes.}
\label{fig:3D-mobility-sequence-diagram}
\end{figure}

Fig.~\ref{fig:3d-mobility-holonic-architecture} illustrates the realisation of the scenario in our proposed architecture. A detailed sequence diagram is shown in Fig.~\ref{fig:3D-mobility-sequence-diagram}.

\subsubsection{Specialized Holons:}
We represent the four specialised holons as follows:
\begin{enumerate}
\item \textit{Resource Holons}: Model the capabilities of resources such as human operators, ground vehicles, aerial vehicles, and passengers. 
\item \textit{Plan Holons}: Responsible for planning passenger trips and routing vehicles. They also design journeys by considering available resources and their current status to ensure optimal paths.
\item \textit{Task Holons}: Select and manage trip segments by scheduling machine resource holons, such as ground or aerial vehicles. They also oversee specific journey segments, such as flying over streets and adapting based on real-time sensing data.
\item \textit{Supervise Holons}: Oversee overall system coordination, emergency response management, resource allocation, and load balancing. They also ensure the timing and interaction between driving and flying tasks and maintain operational efficiency.
\end{enumerate}

\subsubsection{Input:}
The customer communicates their destination to the 3D Mobility SoS by requesting transportation from point \textit{X} to point \textit{Y}.
In this context, the customer is represented as a \textit{human resource holon}, labelled \textit{c1}.
The \textit{c1} can use the natural language capabilities of its LLM to communicate the customer's request to the SoS. 
This holon interfaces with the SoS to make requests, track status updates, and receive progress reports.

\subsubsection{Processing:}
The transportation request from \textit{c1} is received by the \textit{S-SoS}, which serves as the \textit{supervisor holon} of the SoS and functions as the primary coordinating entity.
The \textit{S-SoS} utilises its LLM to refine the input, ensuring it is tailored to the 3D mobility domain. Subsequently, the LLM initiates mission planning.


Upon receiving the request, the \textit{S-SoS} communicates with two other supervisor holons, \textit{S-CS1} and \textit{S-CS2}, each overseeing a different CS of the SoS.
These supervisors collaborate to negotiate and schedule a transportation solution, considering the current status and ongoing tasks of their respective CS.

\subsubsection{Output:}
\label{sec:resulting-uvf}
The supervisor holons collaborate to develop an overarching plan, $Pa$, to fulfil the \textit{c1}'s transportation request.
Plan \(P_a\) consists of three sequential \textit{task holons}: \(T_{a1}\) (driving), \(T_{a2}\) (flying), and \(T_{a3}\) (driving), covering the journey across multiple transportation modes.

Each task holon can be further divided into sub-plans. For example, \(T_{a1}\), representing the initial driving phase, generates a sub-plan, \(P_{a1}\), which specifies detailed navigation instructions similar to those provided by Google Maps. This sub-plan, \(P_{a1}\), is then divided into smaller task holons, such as \(T_{a11}\) and \(T_{a12}\), representing specific navigation actions (e.g., ``Drive along Street X for 50 meters''). These sub-task holons are assigned to and control the \textit{machine resource holons} responsible for this portion of the journey, such as \(m1\) (an autonomous car). Since the customer remains physically bound to this vehicle, both \(T_{a11}\) and \(T_{a12}\) utilise the same resource holon.

This process is similarly repeated for the air-taxi phase \(T_{a2}\), which forms its sub-plan \(P_{a2}\) to execute the flying segment, followed by \(P_{a3}\) for the final driving phase.

\section{Discussion}
\label{sec:discussion}


Our proposed layered architecture for holons differs from the current state-of-the-art, which typically represents holons as CS or abstract IoT devices~\cite{elhabbash2024principled}.
Our reasoning layer provides consistent reasoning capabilities, enabling each holon to adapt autonomously to environmental changes while maintaining coherent behaviour across the system.
Specialised holons are inspired by smart manufacturing~\cite{leitao2006adacor}. Our work is the first to introduce them within the SoS context.

Integrating LLM into the reasoning layer enhances the functionality of each type of holon.
Human resource holon leverages the LLM's natural language capabilities to enable interactions between customers and vehicles, enhancing human-system interaction.
Machine resource holon uses LLM to intelligently process sensor data and allocate dynamic resources, including autonomous vehicles, drones, and other automated systems. 
Plan holons utilise LLMs to analyse user requests and preferences for personalised trip planning.
Additionally, LLMs support the dynamic adjustment of plans based on real-time system status.
Task holons employ LLMs for real-time task status updates, predictive data analysis, and clear communication of task progress.
The supervise holon can conduct LLM-driven analysis of patterns and trends across the SoS to optimise resource allocation, process emergencies intelligently, and generate effective response plans.

The dual-level planning approach in our proposed architecture enhances real-time adaptability and control. For example, in Section~\ref{sec:resulting-uvf}, the high-level plan ($P_a$) considers the overall SoS state and anticipates future needs, while the lower-level plans (e.g., \(P_{a1}\) and \(P_{a2}\)) allow for dynamic, real-time adjustments. By assigning specific task holons to manage detailed steps locally, the system can swiftly adapt to unexpected changes, thereby ensuring responsive adjustments to real-world conditions.
Moreover, this architecture supports resource reallocation, balancing the workload across the network. For instance, while \(P_{a1}\) is assigned to vehicle \(m1\), \(P_{a3}\) is executed by vehicle \(m3\), demonstrating the flexibility of resource switching to distribute tasks efficiently and maintain optimal workflow within the transportation network.

\section{Proposed Evaluation}
\label{sec:proposed-evaluation}
This study presents a conceptual framework, and we recognise the importance of evaluating its effectiveness through experimentation. Future work will involve implementing this case study for evaluation in a simulated environment, similar to the approach used by Sadik~et~al.\cite{sadik2023self}.

\subsection{Experimental Setup}
Simulations can be conducted using multi-agent frameworks, such as JADE, or within a multi-robot environment, like ROS2 and Gazebo, to model interactions.
Additionally, the system can be implemented using a distributed computing framework, with each holon running independently. 
The reasoning layer components can be developed using a fine-tuned version of GPT-3 optimised for transportation-related tasks and interactions.

\subsection{Evaluation Metrics}
We propose the following metrics to evaluate the architecture and compare it with other architectures that do not utilise LLMs and specialised holons (e.g. \cite{elhabbash2024principled,sadik2023self}).

\begin{itemize}
\item \textit{Scalability}: The system's ability to maintain performance as the number of vehicles and users increased.

\item \textit{adaptability}: The system's responsiveness to unexpected events, such as road closures, weather changes, and downtime.

\item \textit{Resource Utilisation}: The average idle time of vehicles and the efficiency of route planning.

\item \textit{Response Time}: The latency in the system's responses to user queries and operational changes.

\item \textit{User Satisfaction}: Evaluated through simulated user interactions and feedback.

\end{itemize}

\section{Conclusion and Future Work}
\label{sec:conclusion-and-future-work}

This study expands the holonic architecture of SoS to tackle adaptability and human interaction challenges.
We have extended the holons by adding three layers, including a reasoning layer that utilises LLM capabilities to perform specific tasks and coordinate with other holons.
To ensure effective  functioning of the SoS, we introduced specialised holons--supervisor, planner, task, and resource--and defined their workflows.
We conceptually demonstrated the proposed architecture's applicability through a 3D mobility case study. This case study illustrates how various autonomous ground and aerial vehicles coordinate effectively.

Future work will implement this case study and compare the results with state-of-the-art approaches. Additionally, we will address the challenges of ambiguity in natural language interactions by incorporating clarification dialogues and improving control mechanisms.
Implementing rigorous verification and validation processes is also essential to ensuring that the LLM's output is robust and reliable.
Finally, future research should also consider potential ethical concerns, including privacy, safety, and conflicts of interest~\cite{rousi2023business,levinson2024privacy}.

\bibliographystyle{splncs04}
\bibliography{main}

\begin{thebibliography}{10}
\providecommand{\url}[1]{\texttt{#1}}
\providecommand{\urlprefix}{URL }
\providecommand{\doi}[1]{https://doi.org/#1}

\bibitem{abbott2006open}
Abbott, R.: Open at the top; open at the bottom; and continually (but slowly) evolving. In: 1st {IEEE/SMC} International Conference on System of Systems Engineering. pp.~1--6. {IEEE} Computer Society (2006)

\bibitem{icsoft24}
Ashfaq, M., Sadik, A.R., Mikkonen, T., Waseem, M., Mäkitalo, N.: Enhancing holonic architecture with natural language processing for system of systems. In: Proceedings of the 19th International Conference on Software Technologies. pp. 427--433. SciTePress (2024)

\bibitem{blair2015holons}
Blair, G.S., Bromberg, Y., Coulson, G., Elkhatib, Y., R{\'{e}}veill{\`{e}}re, L., Ribeiro, H.B., Rivi{\`{e}}re, E., Ta{\"{\i}}ani, F.: Holons: towards a systematic approach to composing systems of systems. In: Proceedings of the 14th International Workshop on Adaptive and Reflective Middleware. pp. 5:1--5:6. {ACM} (2015)

\bibitem{boardman2006system}
Boardman, J.T., Sauser, B.J.: System of systems -- the meaning of \emph{of}. In: 1st {IEEE/SMC} International Conference on System of Systems Engineering. pp. 118--123. {IEEE} Computer Society (2006)

\bibitem{brown2020language}
Brown, T.B., Mann, B., Ryder, N., Subbiah, M., Kaplan, J., Dhariwal, P., Neelakantan, A., Shyam, P., Sastry, G., Askell, A., Agarwal, S., Herbert{-}Voss, A., Krueger, G., Henighan, T., Child, R., Ramesh, A., Ziegler, D.M., Wu, J., Winter, C., Hesse, C., Chen, M., Sigler, E., Litwin, M., Gray, S., Chess, B., Clark, J., Berner, C., McCandlish, S., Radford, A., Sutskever, I., Amodei, D.: Language models are few-shot learners. In: Larochelle, H., Ranzato, M., Hadsell, R., Balcan, M., Lin, H. (eds.) 34th Annual Conference on Neural Information Processing Systems. Advances in Neural Information Processing Systems, vol.~33, pp. 1877--1901. Curran Associates, Inc. (2020)

\bibitem{dahmann2015systems}
Dahmann, J.S.: Systems of systems characterization and types. Systems of Systems Engineering for NATO Defence Applications (STO-EN-SCI-276) pp. 1--14 (2015)

\bibitem{elhabbash2024principled}
Elhabbash, A., Elkhatib, Y., Nundloll, V., Marco, V.S., Blair, G.S.: Principled and automated system of systems composition using an ontological architecture. Future Generation Computer Systems  \textbf{157},  499--515 (2024)

\bibitem{gideon2005taxonomy}
Gideon, J., Dagli, C.H., Miller, A.K.: Taxonomy of systems-of-systems. In: Proceedings of the Conference on Systems Engineering Research. pp. 356--363. IEEE (2005)

\bibitem{jamshidi2008systems}
Jamshidi, M. (ed.): Systems of Systems Engineering: Principles and Applications. CRC Press, 1st edn. (2008)

\bibitem{khurana2023natural}
Khurana, D., Koli, A., Khatter, K., Singh, S.: Natural language processing: state of the art, current trends and challenges. Multimedia tools and applications  \textbf{82}(3),  3713--3744 (2023)

\bibitem{koestler1967ghost}
Koestler, A.: The Ghost in the Machine. Macmillan (1968)

\bibitem{koubaa2024next}
Koubaa, A., Ammar, A., Boulila, W.: Next-generation human-robot interaction with {ChatGPT} and robot operating system. Software: Practice and Experience  (2024)

\bibitem{leitao2006adacor}
Leit{\~a}o, P., Restivo, F.: Adacor: A holonic architecture for agile and adaptive manufacturing control. Computers in industry  \textbf{57}(2),  121--130 (2006)

\bibitem{levinson2024privacy}
Levinson, L., Dietrich, M., Sarkisian, A., Sabanovic, S., Smart, W.D.: Privacy aware robotics. In: Companion of the 2024 ACM/IEEE International Conference on Human-Robot Interaction. pp. 1335--1337. {ACM} (2024)

\bibitem{lopes2016sos}
Lopes, F., Loss, S., Mendes, A., Batista, T., Lea, R.: {SoS-centric} middleware services for interoperability in smart cities systems. In: Proceedings of the 2nd International Workshop on Smart. pp.~1--6. ACM (2016)

\bibitem{maier1998architecting}
Maier, M.W.: Architecting principles for systems-of-systems. Systems Engineering: The Journal of the International Council on Systems Engineering  \textbf{1}(4),  267--284 (1998)

\bibitem{nadira2020towards}
Nadira, B., Bouanaka, C., Bendjaballah, M., Djarri, A.: Towards an {UML-based} {SoS} analysis and design process. In: International Conference on Advanced Aspects of Software Engineering. pp.~1--8. IEEE (2020)

\bibitem{neto2017model}
Neto, V.V.G.: A model-based approach towards the building of trustworthy software-intensive systems-of-systems. In: IEEE/ACM 39th International Conference on Software Engineering Companion. pp. 425--428. IEEE (2017)

\bibitem{nielsen2015systems}
Nielsen, C.B., Larsen, P.G., Fitzgerald, J., Woodcock, J., Peleska, J.: Systems of systems engineering: basic concepts, model-based techniques, and research directions. ACM Computing Surveys  \textbf{48}(2),  1--41 (2015)

\bibitem{nundloll2020ontological}
Nundloll, V., Elkhatib, Y., Elhabbash, A., Blair, G.S.: An ontological framework for opportunistic composition of iot systems. In: IEEE International Conference on Informatics, IoT, and Enabling Technologies. pp. 614--621. IEEE (2020)

\bibitem{radford2019language}
Radford, A., Wu, J., Child, R., Luan, D., Amodei, D., Sutskever, I.: Language models are unsupervised multitask learners. \url{https://cdn.openai.com/better-language-models/language_models_are_unsupervised_multitask_learners.pdf} (2018), last accessed 2024/11/06

\bibitem{rousi2023business}
Rousi, R., Samani, H., M{\"a}kitalo, N., Vakkuri, V., Linkola, S., Kemell, K.K., Daubaris, P., Fronza, I., Mikkonen, T., Abrahamsson, P.: Business and ethical concerns in domestic conversational generative {AI-empowered} multi-robot systems. In: Proceedings of the 14th International Conference on Software Business. Lecture Notes in Business Information Processing, vol.~500, pp. 173--189. Springer (2023)

\bibitem{sadik2023self}
Sadik, A.R., Bolder, B., Subasic, P.: A self-adaptive system of systems architecture to enable its ad-hoc scalability: unmanned vehicle fleet-mission control center case study. In: Proceedings of the 7th International Conference on Intelligent Systems, Metaheuristics \& Swarm Intelligence. pp. 111--118. ACM (2023)

\bibitem{sadik2023analysis}
Sadik, A.R., Ceravola, A., Joublin, F., Patra, J.: Analysis of chatgpt on source code. CoRR  \textbf{abs/2306.00597} (2023)

\bibitem{sadik2017combining}
Sadik, A.R., Urban, B.: Combining adaptive holonic control and isa-95 architectures to self-organize the interaction in a worker-industrial robot cooperative workcell. Future Internet  \textbf{9}(3), ~35 (2017)

\bibitem{silva2020verification}
Silva, E., Batista, T., Oquendo, F.: On the verification of mission-related properties in software-intensive systems-of-systems architectural design. Science of Computer Programming  \textbf{192},  102425 (2020)

\bibitem{song2022continuous}
Song, J., Kang, J., Hyun, S., Jee, E., Bae, D.H.: Continuous verification of system of systems with collaborative {MAPE-K} pattern and probability model slicing. Information and Software Technology  \textbf{147},  106904 (2022)

\bibitem{tchappi2020critical}
Tchappi, I.H., Galland, S., Kamla, V.C., Kamgang, J.C., Mualla, Y., Najjar, A., Hilaire, V.: A critical review of the use of holonic paradigm in traffic and transportation systems. Engineering Applications of Artificial Intelligence  \textbf{90},  103503 (2020)

\bibitem{teixeira2020constituent}
Teixeira, P.G., Lebtag, B.G.A., Dos~Santos, R.P., Fernandes, J., Mohsin, A., Kassab, M., Neto, V.V.G.: Constituent system design: A software architecture approach. In: IEEE International Conference on Software Architecture Companion. pp. 218--225. IEEE (2020)

\bibitem{wudka2020reconfiguration}
Wudka, B., Thomas, C., Siefke, L., Sommer, V.: A reconfiguration approach for open adaptive systems-of-systems. In: IEEE International Symposium on Software Reliability Engineering Workshops. pp. 219--222. IEEE (2020)

\bibitem{zhang2023nlp}
Zhang, Z., Elkhatib, Y., Elhabbash, A.: {NLP-based} generation of ontological system descriptions for composition of smart home devices. In: IEEE International Conference on Web Services. pp. 360--370. IEEE (2023)

\bibitem{zhao2023survey}
Zhao, W.X., Zhou, K., Li, J., Tang, T., Wang, X., Hou, Y., Min, Y., Zhang, B., Zhang, J., Dong, Z., Du, Y., Yang, C., Chen, Y., Chen, Z., Jiang, J., Ren, R., Li, Y., Tang, X., Liu, Z., Liu, P., Nie, J., Wen, J.: A survey of large language models. CoRR  \textbf{abs/2303.18223} (2023)

\end{thebibliography}

\end{document}